\documentclass[10pt,twocolumn,letterpaper]{article}
\usepackage{iccv}
\usepackage{times}
\usepackage{tikz}
\usepackage{pgfplots}
\usepackage{calc}
\usepackage{graphicx}
\usepackage{float}
\usepackage{amsmath}
\usepackage{amssymb}
\usepackage{bm}
\usepackage{booktabs}
\usepackage[format=hang]{caption}       
\usepackage[subrefformat=parens]{subcaption}   

\usepackage{calc}
\usepackage{mathtools}
\usepackage{bm}

\usepackage[short labels]{enumitem}

\usepackage[UKenglish]{babel}
\usepackage[style = british]{csquotes}
\usepackage[per-mode = symbol, exponent-product = \cdot]{siunitx}

\pgfplotsset{compat = newest}
\usetikzlibrary{matrix, positioning}
\usepgfplotslibrary{groupplots, fillbetween, statistics}
\tikzset{>= stealth, every picture/.style = {font=\small}}
\pgfplotsset{
  every tick/.style = {black},
  legend cell align = {left},
  every axis plot/.append style = {thick}
}

\usepackage{colortbl}
\definecolor{myblue}{HTML}{1e2359}
\definecolor{myred}{HTML}{b8261e}
\colorlet{mygreen}{green!60!black}

\usepackage{xspace}

\makeatletter
\DeclareRobustCommand\onedot{\futurelet\@let@token\@onedot}
\def\@onedot{\ifx\@let@token.\else.\null\fi\xspace}

\def\eg{e.g\onedot} 
\def\ie{i.e\onedot}

\def\etal{et al\onedot}
\makeatother

\DeclareMathOperator{\E}{\mathbb{E}}

\DeclarePairedDelimiterX\cond[2]{(}{)}{#1\,\delimsize\vert\,\mathopen{}#2}
\DeclarePairedDelimiterX\expcond[2]{[}{]}{#1\,\delimsize\vert\,\mathopen{}#2}
\DeclarePairedDelimiterX\varcond[2]{.}{.}{#1\,\delimsize\vert\,\mathopen{}#2}
\DeclarePairedDelimiterX\divergence[2]{(}{)}{#1\,\delimsize\Vert\,\mathopen{}#2}

\newcommand{\euler}{\mathrm{e}}
\newcommand{\dd}{\mathrm{d}}
\renewcommand{\vec}[1]{\bm{#1}}
\newcommand{\matr}[1]{\bm{#1}}

\newcommand{\transpose}{\mathsf{T}}

\newcommand{\Gaussian}{\mathcal{N}}

\newcommand{\GP}{\mathcal{GP}}

\newcommand{\independent}{\protect\mathpalette{\protect\independenT}{\perp}}
\def\independenT#1#2{\mathrel{\rlap{$#1#2$}\mkern2mu{#1#2}}}

\newcommand{\bigO}{\mathcal{O}}

\newcommand{\ac}[1]{\textsc{\lowercase{#1}}}

\DeclarePairedDelimiter\rowvect{[}{]}
\DeclarePairedDelimiter\abs{\lvert}{\rvert}

\DeclareMathOperator{\Tr}{Tr}

\newlength{\listindent}
\setlist{leftmargin = \listindent, itemsep = 0.25\baselineskip, topsep = 0.5\baselineskip}

\newenvironment{example}[1][Example]{
  \begin{description}[font = \normalfont \itshape, topsep = \baselineskip, leftmargin = \listindent]
  \item[#1 \quad]
  }{
  \end{description}
}

\usepackage[breaklinks=true,letterpaper=true,colorlinks,bookmarks=false]{hyperref}

\iccvfinalcopy 

\def\iccvPaperID{****} 

\ificcvfinal\fi

\makeatletter
\let\cite\relax
\DeclareRobustCommand{\cite}{%
  \let\new@cite@pre\@gobble
  \@ifnextchar[\new@cite{\@citex[]}}
\def\new@cite[#1]{\@ifnextchar[{\new@citea{#1}}{\@citex[#1]}}
\def\new@citea#1{\def\new@cite@pre{#1}\@citex}
\def\@cite#1#2{[{\new@cite@pre\space#1\if\relax\detokenize{#2}\relax\else, #2\fi}]}
\makeatother

\begin{document}

\title{Estimation of Bivariate Structural Causal Models by\\
  Variational Gaussian Process Regression \\
  Under Likelihoods Parametrised by Normalising Flows
}

\author{Nico Reick, Felix Wiewel, Alexander Bartler and Bin Yang\\
Institute of Signal Processing and System Theory, University of Stuttgart, Germany\\
{\tt\small nico.reick@gmx.de, \{felix.wiewel, alexander.bartler, bin.yang\}@iss.uni-stuttgart.de} \\
}

\maketitle
\ificcvfinal\thispagestyle{empty}\fi

\begin{abstract}
  One major drawback of state-of-the-art artificial intelligence is its lack of explainability.  One approach to solve the problem is taking causality into account.  Causal mechanisms can be described by structural causal models.  In this work, we propose a method for estimating bivariate structural causal models using a combination of normalising flows applied to density estimation and variational Gaussian process regression for post-nonlinear models.  It facilitates causal discovery, \ie distinguishing cause and effect, by either the independence of cause and residual or a likelihood ratio test.  Our method which estimates post-nonlinear models can better explain a variety of real-world cause-effect pairs than a simple additive noise model.  Though it remains difficult to exploit this benefit regarding \emph{all} pairs from the Tübingen benchmark database, we demonstrate that combining the additive noise model approach with our method significantly enhances causal discovery.
\end{abstract}

\section{Introduction}

The success of machine learning and especially of deep neural networks is obvious and manifold.  However, most deep learning algorithms model the underlying system as a \enquote{black box}---they lack explainability.  According to the Turing award winner Judea Pearl \cite{pearlTheoreticalImpedimentsMachine2018}, the problem of state-of-the-art artificial intelligence is being \enquote{model-blind} which intrinsically limits its performance on cognitively more demanding tasks.  Instead of only considering the observational distributions which current deep learning approaches replicate, he proposes to learn underlying causal mechanisms based on structural causal models (\ac{SCM}).

\paragraph{Causal Modelling and Causal Discovery}
There are recent influential works which take up this idea, for instance \cite{pawlowskiDeepStructuralCausal2020} in which a generative model is set up to replicate the generation of \ac{MNIST} digits by a simple causal model: the thickness a digit is drawn with, the intensity of the respective image and their influences on the final image are taken into account.  Having a, in this case at least rudimentary, causal understanding of the generation process, questions like \enquote{How would the image have looked like if the thickness of drawing had been increased?} can be answered, known as counterfactuals.  In this example, the causal relations are designed by a human which is of course undesired.  It would be much more desirable if the underlying causal mechanisms could be inferred from pure data without any expert knowledge.

This problem is addressed by \emph{causal discovery}.  In our work, we restrict ourselves to bivariate structural causal models which are also known as \enquote{cause-effect models}.  It is the simplest case which can be seen as a base of multivariate causal graphs which model real-world data.  Such a structural causal model \( \mathfrak{C} \) is defined by
\begin{align}
  x &\sim p_{x}(x), \\
  \label{eq:general-model}
  y &= f_{y}(x, \varepsilon_{y}) \quad \text{with} \quad \varepsilon_{y} \sim p_{\varepsilon_{y}}(\varepsilon_{y}) \;\; \text{and} \;\; x \independent \varepsilon_{y},
\end{align}
where the common notation \( x \independent \varepsilon_{y} \) is used to denote the independence of the random variables \( x \) and \( \varepsilon_{y} \).

\paragraph{Identifiability}
Unfortunately, only having access to samples of the observational distribution \( p(x, y) \) is not sufficient to distinguish cause from effect because another structural causal model \( \tilde{\mathfrak{C}} \),
\begin{align}
  y &\sim p_{y}(y), \\
  x &= f_{x}(y, \varepsilon_{x}) \quad \text{with} \quad \varepsilon_{x} \sim p_{\varepsilon_{x}}(\varepsilon_{x}) \;\; \text{and} \;\; y \independent \varepsilon_{x},
\end{align}
exists which induces exactly the same observational distribution \cite[\eg\footnote{Comprehensive introductions into causal reasoning are the textbooks \cite{petersElementsCausalInference2017} and \cite{pearlCausalInferenceStatistics2016}, Pearl's monograph \cite{pearlCausalityModelsReasoning2009a} can be considered for a detailed treatise of the subject.  As an extensive review on the estimation of bivariate structural causal models we suggest to refer to \cite{mooijDistinguishingCauseEffect2016}.}][]{petersElementsCausalInference2017}.  What the general structural causal model lacks is \emph{identifiability}:  Only observing samples of \( x \) and \( y \) is not sufficient to distinguish \( x \to y \) from \( y \to x \) since both explanations are possible.  But by restricting the functional form of \( f_{y} \) (and \( f_{x} \) for the non-causal \ac{SCM}) in an appropriate way, the correct causal mechanism can still be identified.  The most common restriction is the (nonlinear) \emph{additive noise model} \cite{hoyerNonlinearCausalDiscovery2009} in which the effect is given by
\begin{equation}
  \label{eq:additive-model}
  y = f(x) + \varepsilon_{y}.
\end{equation}
Furthermore, we consider the \emph{post-nonlinear model} \cite{zhangIdentifiabilityPostNonlinearCausal2009} as an extension which introduces an additional (nonlinear) invertible function \( g \) as follows:
\begin{equation}
  \label{eq:post-nonlinear-model}
  y = g\bigl( \underbrace{f(x) + \varepsilon_{y}}_{\eqqcolon s} \bigr). 
\end{equation}
Both classes of models are identifiable except for some special cases \cite{zhangIdentifiabilityPostNonlinearCausal2009}.  One sufficient condition for a backward post-nonlinear model to not exist, for example, is \( f \) being non-invertible.

\paragraph{Model Selection Criteria}
There are several methods for identifying cause and effect given the previously stated models.  Most of them build upon the \emph{Hilbert-Schmidt Independence Criterion} \cite{grettonKernelStatisticalTest2007}:  As \( x \independent \varepsilon_{y} \) holds by definition, the estimated residuals
\begin{equation}
  \label{eq:residuals}
  \hat{r}_{y} = \hat{g}^{-1}(y) - \E \expcond[\big]{\hat{f}(x)}{x}
\end{equation}
need to be independent of \( x \) if \( p(x, y) \) follows a forward post-nonlinear model\footnote{Setting \( g(s) = s \) gives the residuals of the additive noise model.  In the following, we always discuss the results for the more general post-nonlinear model.} \cite{hoyerNonlinearCausalDiscovery2009, zhangIdentifiabilityPostNonlinearCausal2009}.  The estimates \( \hat{f} \) and \( \hat{g} \) are gained by appropriately regressing \( y \) on \( x \).

One straightforward alternative to the independence test is using the likelihood of \( \mathfrak{C} \) and \( \tilde{\mathfrak{C}} \) as a model selection criterion, \ie following the maximum likelihood principle.  Given a dataset \( \bigl\{ (x_{i}, y_{i}) \bigr\}_{i=1}^{N} \) of \( N \) mutually independent pairs \( (x_{i}, y_{i}) \), we can compare the log-likelihoods
\begin{equation}
  \label{eq:L_xy}
  L_{x \to y} = \sum_{i=1}^{N} \log \hat{p}^{\mathfrak{C}} \cond{y_{i}}{x_{i}; \vec{\theta}} + \sum_{i=1}^{N} \log \hat{p} (x_{i}; \vec{\phi})
\end{equation}
and
\begin{equation}
  \label{eq:L_yx}
  L_{y \to x} = \sum_{i=1}^{N} \log \hat{p}^{\tilde{\mathfrak{C}}} \cond{x_{i}}{y_{i}; \tilde{\vec{\theta}}} + \sum_{i=1}^{N} \log \hat{p} (y_{i}; \tilde{\vec{\phi}})
\end{equation}
with each other and decide for \( x \to y \) to be the causal mechanism if \( L_{x \to y} > L_{y \to x} \) and vice versa otherwise.  \( \vec{\theta} \), \( \tilde{\vec{\theta}} \), \( \vec{\phi} \) and \( \tilde{\vec{\phi}} \) are those parameters maximising the respective likelihoods.  Doing model selection based on the likelihoods is typically infeasible since most expressive models only enable sampling from the density \( \hat{p} \cond{y}{x} \) but not evaluating it.  However, our estimation method provides a lower bound of \( \hat{p} \cond{y}{x} \) and thus enables (approximate) likelihood based causal discovery.

\paragraph{Previous Work}
A variety of methods to discover cause and effect from pure observational data have been proposed.  Many of these are compared in the extensive review \cite{mooijDistinguishingCauseEffect2016} in which, additionally, new benchmark datasets are introduced for the task of causal discovery in the bivariate case.

Hoyer~\etal, who originally proposed the additive noise model, use standard Gaussian process regression to estimate \( \hat{f} \) in a non-parametric way.  This means, the additive noise \( \varepsilon_{y} \) is intrinsically assumed to be Gaussian \cite{hoyerNonlinearCausalDiscovery2009}.  Mooij~\etal slightly adapt the regression in their review \cite{mooijDistinguishingCauseEffect2016}.  They use \ac{FITC} \cite{quinonero-candelaUnifyingViewSparse2005} as an approximation to exact Gaussian process regression in favour of a reduced computational complexity.  The Gaussianity assumption is maintained.

Most influential for our work is \cite{zhangEstimationFunctionalCausal2016} in which post-nonlinear models are studied.  The authors claim and illustrate that a wrong assumption on \( p_{\varepsilon_{y}} \) would lead to a biased estimate of \( f \) and \( g \).  Therefore, they allow for a Gaussian mixture of the noise \( \varepsilon_{y} \).  While \( \hat{f} \) is again the mean of a posterior Gaussian process, the inverse of \( g \) is a member of a parametric family of functions defined by
\begin{equation}
  g^{-1}(y; \vec{a}, \vec{b}, \vec{c}) = y + \sum_{i=1}^{k} a_{i} \tanh \bigl( b_{i}(y + c_{i}) \bigr),
\end{equation}
where \( k \) is a hyperparameter and \( \vec{a} \), \( \vec{b} \), \( \vec{c} \) are all contained in \( \vec{\theta} \).
Because of the non-Gaussian noise \( \varepsilon_{y} \) the likelihood is not Gaussian anymore.  Zhang~\etal use Monte Carlo Expectation Maximisation (\ac{EM}) to learn the parameters of \( g^{-1} \), \( p_{\varepsilon_{y}} \) and of the process's kernel \cite{zhangEstimationFunctionalCausal2016}.

\paragraph{Our Contribution}
We present an alternative to this method, which uses variational inference and a noise distribution parametrised by normalising flows \cite{papamakariosNormalizingFlowsProbabilistic2019, durkanNeuralSplineFlows2019}.  We therefore adapt the method \cite{hensmanScalableVariationalGaussian2015} which was originally proposed in the field of scalable Gaussian process classification.  The main contributions of our work can be summarised under the following points:
\begin{itemize}
\item We propose a Gaussian regression method for data which follow post-nonlinear models.  We use normalising flows to model an arbitrary noise distribution and combine these with a variational Gaussian process regression approach \cite{hensmanScalableVariationalGaussian2015}.  Our method is considerably less time-consuming than the approach in \cite{zhangEstimationFunctionalCausal2016}.  It reduces the computation time from more than ten hours to less than one minute.

\item Moreover, we show how the method can be applied to estimate bivariate structural causal models and demonstrate how causal discovery based on the marginal likelihood (more precisely a lower bound of the log marginal likelihood) or well-known independence tests can be performed.
  
\item We finally compare the usage of classical discovery methods based on the simple additive noise model to our approach on the Tübingen Database with Cause-Effect Pairs \cite{mooijDistinguishingCauseEffect2016} and present a heuristic method which combines both approaches.
\end{itemize}

\section{Estimation of Bivariate \ac{SCM}s}

In this section, we propose a novel regression method to estimate causal mechanisms which can be represented by bivariate post-nonlinear \ac{SCM}s
\begin{equation}
  y = g \bigl( f(x) + \varepsilon_{y} \bigr).
\end{equation}

Our discussion starts with normalising flows which provide a framework to approximate complicated noise distributions \( \varepsilon_{y} \) while still enabling to evaluate the estimated density of \( \varepsilon_{y} \).

\subsection{Normalising Flows}
\label{sec:normalising-flows}

Most of the preliminary works on bivariate \ac{SCM}s as \cite{hoyerNonlinearCausalDiscovery2009} and \cite[Section~4.1]{mooijDistinguishingCauseEffect2016} assume the additive noise to be Gaussian distributed, motivated by the central limit theorem.  However, Gaussian noise is a strong assumption not always fulfilled in practice\footnote{It is for example well known from a variety of engineering problems, as telecommunications, image processing, underwater acoustics, radar signal processing etc., that the densities of the noise variables are often heavy tailed \cite{nolanUnivariateStableDistributions2020}; not Gaussian but so called \emph{impulsive noise} is corrupting the respective signals.}.  By transforming a Gaussian distributed random variable in an appropriate way, various non-Gaussian noise distributions can be realised.

One method to estimate the noise distribution from pure data is the concept of \emph{normalising flows} \cite{papamakariosNormalizingFlowsProbabilistic2019}.  Let \( x \) be a continuous random variable with density \( p_{x}(x) \).  \( x \) is a transformation \( T \) of a random variable \( u \) distributed according to the \emph{base distribution} \( p_{u}(u) \).  Both \( T \) and \( p_{u} \) can be parametric, i.e. \( T(u) = T(u; \vec{\chi}) \) and \( p_{u}(u) = p_{u} (u; \vec{\psi}) \) with the parameter vectors \( \vec{\chi} \) and \( \vec{\psi} \).  Typical parameters of the base distribution are the mean \( \mu \) and variance \( \sigma^{2} \) if we choose a normal base distribution as we will do in the following\footnote{We force the mean of the Gaussian base distribution to zero.}.  In flow-based models, \( T \) is a diffeomorphism\footnote{\ie \( T \) is bijective, \( T \) and its inverse \( T^{-1} \) are differentiable.}.  It is possible to sample from the model by only sampling from the simple base distribution and applying \( T \).  Furthermore, we can analytically evaluate our approximation \( q_{x}(x) \) of the density \( p_{x}(x) \) according to
\begin{equation}
  \label{eq:change-of-variables}
  q_{x}(x) = p_{u} \bigl( T^{-1}(x) \bigr) \abs*{\frac{\dd T^{-1}(x)}{\dd x}}
\end{equation}
if \( T^{-1} \) and its derivative are known.

This suffices to fit \( q_{x}(x) \) to given samples of \( x \) using maximum likelihood.  There are several appropriate families of transformations \( T \).  In particular \emph{linear rational spline flows} \cite{dolatabadiInvertibleGenerativeModeling2020} have proven useful during our investigations.  Those define \( T \) piecewise:  \( K \) bins of varying width and height span the region \( -B \leq u \leq B \) and \( -B \leq x \leq B \).  The number of bins \( K \) and the boundary \( B \) are treated as hyperparameters.  Within each bin, a monotonically increasing rational function
\begin{equation}
  T(u) = \frac{au + b}{cu + d}
\end{equation}
is defined.  The coefficients \( \{ a, b, c, d \} \) can be directly calculated from the parameter vector \( \vec{\phi} \) of the transform.  \( \vec{\phi} \) contains the bin widths and heights, the derivatives at the knots (\ie the intersections of neighbouring bins) and one additional parameter per bin which is necessary to obtain unique solutions for the coefficients of the rational function.  See \cite[Algorithm~1]{dolatabadiInvertibleGenerativeModeling2020} for more details.

The estimation of the parameters can be performed iteratively by stochastic gradient descent or, for faster convergence, by Adam \cite{kingmaAdamMethodStochastic2014} or similar advanced stochastic optimisers.  Though constructed for the use with high-dimensional data and in combination with other stacked transformations, one rational spline transformation as defined above is sufficiently expressive to (almost perfectly) replicate one-dimensional densities, even if these are multimodal as we will demonstrate in Section~\ref{sec:artificial-data}.

\subsection{Gaussian Processes}
\label{sec:gaussian-processes}

As an example of parametric modelling, a neural network can be used to approximate the nonlinear function \( f \).  However, it is difficult to choose the \enquote{complexity} of such models appropriately:  Under- or overestimating the required number of layers and neurons per layer results in under- or overfitting, respectively.  Gaussian processes, on the contrary, are non-parametric and as all Bayesian methods, the additional prior information enables an intrinsic \enquote{smoothing}.  Especially on small datasets, where the higher computational complexity can be afforded, Gaussian process regression is known to perform better than regression based on neural networks.

We therefore model the nonlinear function \( f \) by the mean function of a Gaussian process \cite{rasmussenGaussianProcessesMachine2006}.  First, an important note regarding our notation:  The function \( f \) forming the additive model \eqref{eq:additive-model} was considered deterministic.  When modelling \( f \) by a Gaussian process, however, the values \( f_{i} \) which correspond to the given locations \( x_{i} \), i.e. \( f_{i} = f(x_{i}) \), are \emph{not} deterministic anymore.  Moreover, we do not think of \( f(x_{i}) \) as a deterministic transform of some random \( x_{i} \).  We rather place a Gaussian process prior on \( f \) which results in \( \rowvect{f_{i}, \dotsc, f_{N}}^{\transpose} \) being jointly Gaussian distributed.  \( x_{i} \) additionally follows some distribution which is of less importance at this point as we only consider distributions conditioned on \( x_{i} \).  The deterministic \( f \) addressed in \eqref{eq:additive-model} is estimated by the mean of the posterior.  Though also denoting this function by \( f \) is slightly abusive, we are convinced that this does not impair the understanding.

For the prior \( f \sim \GP\bigl( m(x), k(x, x') \bigr) \), we choose \( m(x) \equiv 0 \) and the squared exponential kernel\footnote{We have also experimented with members of the Mat\'ern family because (the more popular) squared exponential kernels are known to lead to very smooth processes which rarely occur in physics
\cite{steinInterpolationSpatialData1999} (as cited in \cite{rasmussenGaussianProcessesMachine2006}).  However, we could not observe a significant benefit from these kernels, so we left it at the squared exponential kernels.}
\begin{align}
  \label{eq:kernel}
  k(x, x') = \exp \left( -\frac{1}{2l^{2}} (x - x')^{2} \right).
\end{align}
If we restricted ourselves to Gaussian noise \( \varepsilon_{y} \sim \Gaussian(0, \sigma_{\varepsilon}^{2}) \), the log marginal likelihood
\begin{align}
  \label{eq:log-marginal-likelihood}
  \log p \cond{\vec{y}}{\vec{x}; \vec{\theta}}
  \notag
  &= -\frac{N}{2} \log (2\pi) - \frac{1}{2} \log\,\abs*{ \matr{K}_{\vec{x}, \vec{x}} + \sigma_{\varepsilon}^{2} \matr{I} } \\
  & \qquad- \frac{1}{2} \vec{y}^{\transpose} \left[ \matr{K}_{\vec{x}, \vec{x}} + \sigma_{\varepsilon}^{2} \matr{I} \right]^{-1} \vec{y}
\end{align}
would be given by an analytical expression, where we subsumed \( l \) and \( \sigma^{2}_{\varepsilon} \) under the parameter vector \( \vec{\theta} = [l, \sigma_{\varepsilon}^{2}]^{\transpose} \).  In the above expression, \( \vec{x} \) and \( \vec{y} \) contain all observed data, \( \matr{K}_{\vec{x}, \vec{x}'} \) denotes the Gram matrix with respect to the kernel \( k(x, x') \).  Equation~\eqref{eq:log-marginal-likelihood} could be maximised to determine \( \vec{\theta} \) which gives the best fit of the process to the data.

\subsection{Variational Gaussian Process Regression}
\label{sec:vari-gauss-proc}

However, in general we intend to model non-Gaussian additive noise.  Motivated by applications in classification instead of regression, Hensman~\etal proposed a variational lower bound on the aforementioned log marginal likelihood in equation~\eqref{eq:log-marginal-likelihood} \cite{hensmanScalableVariationalGaussian2015}.  We can directly apply their results when modelling \( \varepsilon_{y} \) by normalising flows.  For a better understanding, we shortly review \cite{hensmanScalableVariationalGaussian2015}'s work.  For sake of ease, we omit to explicitly state that the following densities are conditioned on the input locations \( \vec{x} \).  Moreover, we first neglect the nonlinearity \( g \).  In Section~\ref{sec:extens-post-nonl} we will discuss how to easily adapt the procedure to allow for estimating \( g \) hand in hand with \( f \) and \( \varepsilon_{y} \).
\begin{itemize}
\item Inference based on exact Gaussian process regression scales with \( \bigO(N^{3}) \).  Sparse approximate methods \cite{quinonero-candelaUnifyingViewSparse2005} introduce \( M \ll N \) additional latent variables, called \emph{inducing variables}.  They are stacked into \( \vec{u} \) and \enquote{compress} the information contained in \( \vec{f} \), the (unnoisy) function values of the Gaussian process at \( \vec{x} \).  \( \vec{u} \) is in a similar way related to the inducing input points \( \vec{z} \) which are learned when maximising the lower bound on the marginal likelihood later on.
  
\item Further bounding the evidence lower bound (\ac{ELBO}), the authors derive
  \begin{align}
  \log p(\vec{y})
  \label{eq:final-lower-bound}
  \notag \geq \mathcal{L} &\coloneqq \sum_{i=1}^{N}\ \E_{q(f_{i})} \bigl[ \log p \cond{y_{i}}{f_{i}} \bigr] \\
  &\qquad - D_{\mathrm{KL}} \divergence[\big]{q(\vec{u})}{p(\vec{u})}
\end{align}
where they introduced the abbreviation
\begin{displaymath}
  q(f_{i}) \coloneqq \int p \cond{f_{i}}{\vec{u}} q(\vec{u}) \,\dd \vec{u}.
\end{displaymath}
\( q(\vec{u}) \), the variational approximation of \( p \cond{\vec{u}}{\vec{y}} \), is chosen to be Gaussian,
\begin{equation}
  q(\vec{u}) = \Gaussian (\vec{u}; \vec{m}, \matr{S}),
\end{equation}
with the mean \( \vec{m} \) and the covariance matrix \( \matr{S} \).  \( q(f_{i}) \) is therefore Gaussian as well and the expectations with respect to it can be numerically determined using the \emph{Gauss-Hermite quadrature} \cite[Section~25.4.46]{abramowitzHandbookMathematicalFunctions1964}.
Furthermore, the second term of the bound, the \ac{KL}~divergence, can be analytically derived:
\begin{align}
    D_{\mathrm{KL}} \divergence[\big]{q(\vec{u})}{p(\vec{u})}
    \notag &= \frac{1}{2} \Bigl( \Tr \bigl( \matr{K}_{\vec{z}, \vec{z}} \matr{S} \bigr) - M \\
    \notag &\qquad\quad + \log\, \abs{\matr{K}_{\vec{z}, \vec{z}}} - \log\, \abs{\matr{S}} \\
    &\qquad\quad + \vec{m}^{\transpose} \matr{K}_{\vec{z}, \vec{z}}^{-1} \vec{m} \Bigr).
\end{align}
\end{itemize}

As we model \( \varepsilon_{y} \) by a spline flow, we have direct access to an analytical expression of
\begin{equation}
  \log p \cond{y_{i}}{f_{i}} = \log p_{\varepsilon_{y}}\bigl( y_{i} - f_{i} \bigr).
\end{equation}
To fit the process, we iteratively maximise \eqref{eq:final-lower-bound} with respect to the following parameters:
\begin{itemize}
\item \( \vec{m} \) and \( \matr{S} \)\footnote{To enable an unconstrained optimisation, the Cholesky decomposition \( \matr{L} \) with \( \matr{S} = \matr{L} \matr{L}^{\transpose} \) is typically chosen as a parameter instead of \( \matr{S} \) itself.} which serve as the mean and covariance matrix of the variational approximation \( q(\vec{u}) \) of \( p \cond{\vec{u}}{\vec{y}} \);
\item \( \vec{z} \), the locations the inducing variables \( \vec{u} \) are associated with;
\item \( \vec{\theta} \) which comprises the parameters of the kernel and the likelihood, i.e. \( l \) in case of the kernel \eqref{eq:kernel}, the parameters \( \sigma^{2} \) and \( \vec{\chi} \) of our base distribution and of the linear rational spline transform, respectively.
\end{itemize}

To replicate data points by the learned model, we draw samples of the posterior predictive distribution
\begin{equation}
  p \cond{\vec{f}_{*}}{\vec{y}}
  \approx \int p \cond{\vec{f}_{*}}{\vec{u}} q(\vec{u}) \,\dd \vec{u}.
\end{equation}
\( \vec{f}_{*} \) denotes the latent function values at (unseen) input locations \( \vec{x}_{*} \).  As \( \vec{f}_{*} \) and \( \vec{u} \) are jointly Gaussian,
\begin{align}
  p \cond{\vec{f}_{*}}{\vec{u}}
  \notag &= \Gaussian \bigl( \vec{f}_{*}; \matr{K}_{\vec{x}_{*}, \vec{z}} \matr{K}_{\vec{z}, \vec{z}}^{-1} \vec{u}, \\
  &\qquad\quad \matr{K}_{\vec{x}_{*}, \vec{x}_{*}} - \matr{K}_{\vec{x}_{*}, \vec{z}} \matr{K}_{\vec{z}, \vec{z}}^{-1} \matr{K}_{\vec{z}, \vec{x}_{*}} \bigr)
\end{align}
can be derived analytically.

\subsection{Extension to Post-Nonlinear Model}
\label{sec:extens-post-nonl}

Adapting the presented regression method such that it can cope with the post-nonlinear model is not demanding:  The fact that \( g \) needs to be an invertible function enables us to parametrise it by another rational spline flow.  The likelihood \( p \cond[\big]{y}{f(x)} \), which plays a major role during regression, is still tractable.  To see this, we first rename the (now latent) variable \( f(x) + \varepsilon_{y} \) to \( s \).  Under this new notation,
\begin{equation}
  \log p_{s} \cond[\big]{s}{f(x)} = \log p_{\varepsilon_{y}} \bigl( s - f(x) \bigr)
\end{equation}
is already known.  With \( g(s) \) being a parametric function \( g_{\vec{\varphi}}(s) \) with its inverse \( s = g_{\vec{\varphi}}^{-1}(y) \), the desired likelihood is given by
\begin{align}
  \log p \cond[\big]{y}{f(x)}
  \notag &= \log p_{\varepsilon_{y}} \Bigl( g_{\vec{\varphi}}^{-1}(y) - f(x) \Bigr) \\
  &\qquad + \log\, \abs*{\frac{\dd g_{\vec{\varphi}}^{-1}(y)}{\dd y}}.
\end{align}

As we again choose linear rational splines to model \( g \), its parameters \( \vec{\varphi} \) contain, among others, the bin widths and heights and the derivatives at the knots.  These modifications are all it takes to expand our regression method to the post-nonlinear model.

\section{Identification of Cause and Effect}
\label{sec:ident-cause-effect}

In the following, we discuss how the presented regression technique can be applied to the initial problem of identifying cause and effect in the bivariate case.

To this end, we regress \( y \) on \( x \) to estimate \( \mathfrak{C} \) and \( x \) on \( y \) to estimate \( \tilde{\mathfrak{C}} \) given a normalised\footnote{\( x \) and \( y \) have zero mean and unit variance.  With normalised datasets, we benefit from hyperparameters of the spline flows that are easier to set, especially the bound \( B \).} dataset \( \bigl\{ (x_{i}, y_{i}) \bigr\}_{i=1}^{N} \).  For this we maximise \eqref{eq:final-lower-bound} with respect to the model's parameters \( \vec{\theta} \) (and the auxiliary variables \( \vec{m} \), \( \matr{S} \), \( \vec{z} \) of the variational approach).  This lower bound is the first contribution to the log-likelihood \( L_{x \to y} \) (or \( L_{y \to x} \) for the backward model).  The other contribution is dependent on the marginal density of \( x \) (or \( y \) for the backward model) which we also estimate by rational spline flows.  Given those, we can select \( \mathfrak{C} \) or \( \tilde{\mathfrak{C}} \) based on maximum likelihood.

After the optimisation, we have access to the posterior predictive density of the Gaussian process.  Its mean is \( \E \expcond[\big]{\hat{f}(x_{i})}{x_{i}} \) which we need to test for the independence of the residuals \( \hat{r}_{y_{i}} \) \eqref{eq:residuals}
and \( x_{i} \) (in the case of \( \mathfrak{C} \); and accordingly \( \hat{r}_{x_{i}} \) and \( y_{i} \) for \( \tilde{\mathfrak{C}} \)).  Checking for independence of \( \bigl\{ (x_{i}, r_{y_{i}}) \bigr\}_{i=1}^{N} \) is done by means of the \ac{HSIC}.  It suffices to only consider the \( p \)-value \cite{hoyerNonlinearCausalDiscovery2009} (or, alternatively, the test statistic \cite{mooijDistinguishingCauseEffect2016}) of the test.  A high \( p \)-value (a low test statistic) corresponds to a high probability of falsely rejecting the independence.  Based on this criterion, we regard that model to be the causal one which leads to a higher \( p \)-value (a lower test statistic) in the hypothesis test.\footnote{Sometimes, the \( p \)-value for both \( \mathfrak{C} \) and \( \tilde{\mathfrak{C}} \) is high and comparable which can be interpreted as follows:  The data fit to post-nonlinear models \( \mathfrak{C} \) \emph{and} \( \tilde{\mathfrak{C}} \).  A possible reasoning might be that the true data generating process meets a non-identifiable post-nonlinear model as listed in \cite{zhangIdentifiabilityPostNonlinearCausal2009}.}

To sum up, our regression method enables causal discovery by means of two criteria:  maximum likelihood and the independence of cause and residuals.  Unfortunately, using the lower bound of the likelihood as the model selection criterion underperforms on practical examples as will be demonstrated in Section~\ref{sec:tubing-datab-with}.

\section{Experiments}
\label{sec:experiments}
The applicability of our estimation approach is now demonstrated on both artificial data and the real-world cause-effect pairs of the Tübingen database.

All of the following simulation results are generated using a squared exponential kernel as given in equation~\eqref{eq:kernel}.  Furthermore, the hyperparameters of all involved rational splines are set to \( K = 21 \) and \( B = 10 \).  We perform \num{1000} steps of stochastic gradient descent.

\subsection{Artificial Data}
\label{sec:artificial-data}

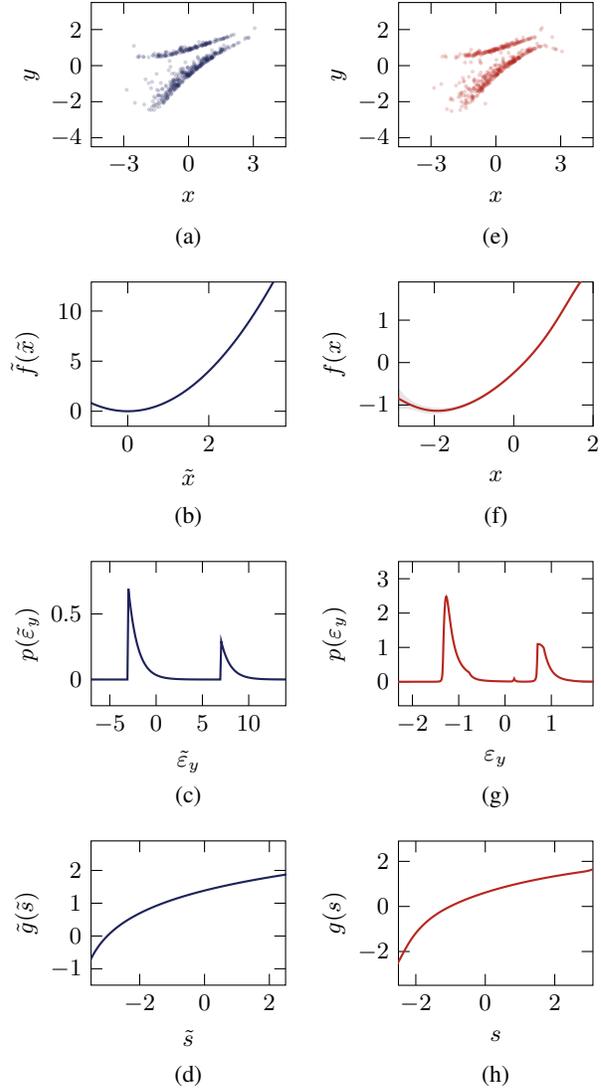
\begin{figure}
  \centering
  \def\labelwidth{10mm}
  \def\labelshift{-11mm}
  \begin{tikzpicture}
    \begin{groupplot}[
      group style = {
        rows = 4, columns = 2,
        horizontal sep = 1.5cm,
        vertical sep = 1.8cm,
      },
      width = 0.5\linewidth, height = 0.42\linewidth,
      ylabel style = {at = {(-0.22, 0.5)}}
      ]

      \nextgroupplot[
      xlabel = \( x \), ylabel = \( y \),
      xtick distance = 3,
      xmin = -4.5, xmax = 4.5, ymin = -4.5, ymax = 3.5]

      \addplot[only marks, mark size=0.75pt, myblue, opacity=0.2, draw opacity=0] table {plot_data/example_dataset.txt};

      \nextgroupplot[
      xtick distance = 3,
      xlabel = \( x \), ylabel = \( y \),
      xmin = -4.5, xmax = 4.5, ymin = -4.5, ymax = 3.5]

      \addplot[only marks, mark size=0.75pt, myred, opacity=0.2, draw opacity=0] table {plot_data/example_dataset_replication.txt};

      \nextgroupplot[
      xlabel = \( \tilde{x} \), ylabel = \( \tilde{f}(\tilde{x}) \),
      xmin = -0.9, xmax = 3.9, ymin = -1.5, ymax = 12.9]

      \addplot[myblue, samples=200] {x^2};

      \nextgroupplot[
      xlabel = \( x \), ylabel = \( f(x) \),
      xmin = -2.9, xmax = 2, ymin = -1.5, ymax = 1.9]

      \addplot[draw = none, name path = upper] table {plot_data/example_pred_lower.txt};
      \addplot[draw = none, name path = lower] table {plot_data/example_pred_upper.txt};
      \addplot[opacity = 0.1] fill between[of = lower and upper];
      
      \addplot[myred] table {plot_data/example_pred_mean.txt};

      \nextgroupplot[
      xlabel = \( \tilde\varepsilon_{y} \), ylabel = \( p(\tilde\varepsilon_{y}) \),
      xmin = -7, xmax = 14, ymin = -0.2, ymax = 0.9
      ]

      \addplot[myblue] table {plot_data/example_noise_pdf.txt};

      \nextgroupplot[
      xlabel = \( \varepsilon_{y} \), ylabel = \( p(\varepsilon_{y}) \),
      xmin = -2.3, xmax = 1.9, ymin = -0.7, ymax = 3.5
      ]

      \addplot[myred] table {plot_data/example_est_noise_pdf.txt};

      \nextgroupplot[
      xlabel = \( \tilde s \), ylabel = \( \tilde g(\tilde s) \),
      xmin = -3.5, xmax = 2.5, ymin = -1.5, ymax = 2.9]
      \addplot[myblue, samples=200] {ln(x + 4)};
      
      \nextgroupplot[
      xlabel = \( s \), ylabel = \( g(s) \),
      xmin = -2.5, xmax = 3.1, ymin = -3.5, ymax = 2.9]
      \addplot[myred, samples=200] table {plot_data/example_post.txt};

    \end{groupplot}

    \node[text width = \labelwidth] at ([yshift=\labelshift] group c1r1.south) {\subcaption{\label{fig:regression-exponential-noise-1}}};
    \node[text width = \labelwidth] at ([yshift=\labelshift] group c1r2.south) {\subcaption{\label{fig:regression-exponential-noise-2}}};
    \node[text width = \labelwidth] at ([yshift=\labelshift] group c1r3.south) {\subcaption{\label{fig:regression-exponential-noise-3}}};
    \node[text width = \labelwidth] at ([yshift=\labelshift] group c1r4.south) {\subcaption{\label{fig:regression-exponential-noise-4}}};
    \node[text width = \labelwidth] at ([yshift=\labelshift] group c2r1.south) {\subcaption{\label{fig:regression-exponential-noise-5}}};
    \node[text width = \labelwidth] at ([yshift=\labelshift] group c2r2.south) {\subcaption{\label{fig:regression-exponential-noise-6}}};
    \node[text width = \labelwidth] at ([yshift=\labelshift] group c2r3.south) {\subcaption{\label{fig:regression-exponential-noise-7}}};
    \node[text width = \labelwidth] at ([yshift=\labelshift] group c2r4.south) {\subcaption{\label{fig:regression-exponential-noise-8}}};
  \end{tikzpicture}

  \caption[Regression on post-nonlinear model with multimodal noise]{Regression on post-nonlinear model with multimodal noise.  The (normalised) data points are highlighted by small transparent dots in panel~\subref{fig:regression-exponential-noise-1};  \subref{fig:regression-exponential-noise-2}--\subref{fig:regression-exponential-noise-4} visualise the (unnormalised) data generating process defined by equations~\eqref{eq:ex-1}--\eqref{eq:ex-4}.  The results when using our variational approach are displayed in  \subref{fig:regression-exponential-noise-5}--\subref{fig:regression-exponential-noise-8}.  These panels show the replication of the dataset (\ie \num{500}~samples drawn from the learned model), posterior predictive distribution (visualised by its mean and confidence interval) the estimated noise distribution and the estimated post-nonlinear function, respectively.

  Remark:  The different scale and shift of the estimated functions is only partly caused by the normalisation of the data.  Because any affine function chained with \( g \) can be fully merged into \( f \) and \( \varepsilon_{y} \) without changing \( p(x, y) \), the original scale cannot be recovered.}
  \label{fig:regression-exponential-noise}
\end{figure}

We start with the following artificial example:  Let \( p(\varepsilon_{y}) \) be a multimodal density comprising two exponential densities:  
\begin{equation}
  \label{eq:ex-1}
  p(\tilde{\varepsilon}_{y}) =
  \begin{dcases}
    0, & \tilde{\varepsilon}_{y} < -3, \\
    0.7\euler^{-(\tilde{\varepsilon}_{y}+3)}, & -3 \leq \tilde{\varepsilon}_{y} < 7, \\
    0.7\euler^{-(\tilde{\varepsilon}_{y}+3)} \\
    \quad+ 0.3 \euler^{-(\tilde{\varepsilon}_{y} - 7)}, & \tilde{\varepsilon}_{y} \geq 7.
  \end{dcases}
\end{equation}
This density corresponds to the unnormalised noise variable \( \tilde{\varepsilon}_{y} \).  We normalise our dataset such that both \( x \) and \( y \) have zero mean and unit variance.  The density \( \tilde{\varepsilon}_{y} \) is shown in Figure~\ref{fig:regression-exponential-noise-3}.

The full process we want to analyse in this example reads
\begin{align}
  \label{eq:ex-2}
  \tilde{x}_{i} &\sim \Gaussian(2, 1), \\
  \label{eq:ex-3}
  \tilde{s}_{i} &= \tilde{x}_{i}^{2} + \tilde{\varepsilon}_{y, i} \quad \text{with} \quad \tilde{\varepsilon}_{y, i} \sim p(\tilde{\varepsilon}_{y}), \\
  \label{eq:ex-4}
  \tilde{y}_{i} &= \log (\tilde{s}_{i} + 4).
\end{align}
We draw \( 500 \)~samples from the model.  The resulting normalised dataset is shown in Figure~\ref{fig:regression-exponential-noise} along with the regression results of our method.  Apparently, it leads to a suitable explanation of the data.

For the simulation, we utilised \( M = 20 \) inducing variables.  Using more inducing variables leads to better results; \( M \lesssim 5 \), on the contrary, is a too high \enquote{compression}.

\subsection{Tübingen Database with Cause-Effect Pairs}
\label{sec:tubing-datab-with}

Whether causal discovery works for real-world problems, is typically assessed by means of a (growing) dataset presented in the article \cite[Appendix~D]{mooijDistinguishingCauseEffect2016}.

We first pick two cause-effect pairs of the database which can be regarded as representative for opportunities of the more general post-nonlinear model compared to the additive noise model, on the one hand, and corresponding difficulties with respect to causal discovery, on the other hand.

\begin{enumerate}[i), nosep, listparindent = \parindent, itemsep = 0.25\baselineskip, topsep = 0.5\baselineskip]
\item \label{item:1} \emph{Causal discovery based on the post-nonlinear model succeeds---causal discovery based on the additive noise model fails:}

  In general, there are a variety of functional forms which \( f_{y} \) as defined in equation~\eqref{eq:general-model} can take.  Whether a simple additive noise model \eqref{eq:additive-model} suffices, is not known a priori.  If so, any more complicated function \( f_{y} \) can be regarded as less plausible according to Occam's Razor \cite{hoyerNonlinearCausalDiscovery2009} (see also \cite[Principle~1]{mooijDistinguishingCauseEffect2016}).

  One cause-effect pair in which neither \( x \to y \) (the true causal direction) nor \( y \to x \) follows an additive noise model, is pair~2 where the altitude (\( x \)) of a weather station of the Deutscher Wetterdienst (\ac{DWD}) and the corresponding mean annual precipitation (\( y \)) were collected.  Figure~\ref{fig:alt_prec} shows that a second nonlinearity (which is monotonically increasing) is required to model the data properly.  The data points drawn from the estimated post-nonlinear model are apparently closer to the actual data than data points drawn from the estimated additive noise model.  We can confirm this observation quantitatively by means of the respective likelihoods.  The lower bound on \( L_{x \to y, \text{\subref{fig:alt_prec-1}}} \) for the post-nonlinear model is approximately \num{-1.72} and thus significantly higher than the respective (unbounded) log-likelihood \( L_{x \to y, \text{\subref{fig:alt_prec-2}}} = -1.93 \).

  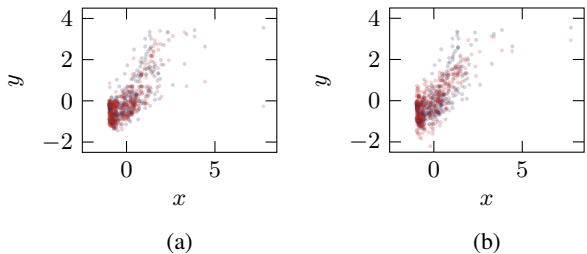
\begin{figure}[t]
    \centering
    \def\labelwidth{10mm}
    \def\labelshift{-11mm}
    \begin{tikzpicture}
      \begin{groupplot}[
        group style = {
          rows = 1, columns = 2,
          horizontal sep = 1.5cm,
          vertical sep = 2cm,
        },
        width = 0.5\linewidth, height = 0.42\linewidth,
        ]

        \nextgroupplot[
        xlabel = \( x \), ylabel = \( y \),
        xmin = -2.5, xmax = 8.5, ymin = -2.5, ymax = 4.5]

        \addplot[only marks, mark size=0.75pt, myblue, opacity=0.2, draw opacity=0] table {plot_data/alt_prec_dataset.txt};
        \addplot[only marks, mark size=0.75pt, myred, opacity=0.2, draw opacity=0] table {plot_data/alt_prec_dataset_replication.txt};
        
        \nextgroupplot[
        xlabel = \( x \), ylabel = \( y \),
        xmin = -2.5, xmax = 8.5, ymin = -2.5, ymax = 4.5]

        \addplot[only marks, mark size=0.75pt, myblue, opacity=0.2, draw opacity=0] table {plot_data/alt_prec_dataset.txt};
        \addplot[only marks, mark size=0.75pt, myred, opacity=0.2, draw opacity=0] table {plot_data/alt_prec_dataset_replication_gauss.txt};
      \end{groupplot}

      \node[text width = \labelwidth] at ([yshift=\labelshift] group c1r1.south) {\subcaption{\label{fig:alt_prec-1}}};
      \node[text width = \labelwidth] at ([yshift=\labelshift] group c2r1.south) {\subcaption{\label{fig:alt_prec-2}}};
    \end{tikzpicture}

    \caption[Investigation of pair~2 (altitude~\( \to \)~precipitation)]{Investigation of pair~2 (altitude~\( \to \)~precipitation): The data can be better explained by allowing for a bijective post-nonlinear function.  The replication of the original dataset (small blue dots) by generating samples from the estimated post-nonlinear model using our proposed method (small red dots in panel~\subref{fig:alt_prec-1}) suits better compared to samples taken from the estimated additive noise model (small red dots in panel~\subref{fig:alt_prec-2}).
    }
    \label{fig:alt_prec}
  \end{figure}

\item \label{item:2} \emph{Causal discovery based on the additive noise model succeeds---causal discovery based on the post-nonlinear model fails:}

  Unfortunately, using the more expressive post-nonlinear model instead of the additive noise model does not always enhance causal discovery based on \ac{HSIC}.  For a significant subset of the cause-effect pairs, the situation is as follows:  there is an additive noise model for \( x \to y \), but the data also fit to post-nonlinear models for both \( x \to y \) and \( y \to x \), with the latter producing a higher \( p \)-value during the independence test.

  One example is pair~4 which is again related to meteorological data from the \ac{DWD}:  \( x \) describes the altitude of the weather station, \( y \) the mean annual sunshine duration.  Figure~\ref{fig:alt_sun} shows that causal discovery based on the post-nonlinear model can fail due to an existing anti-causal (\( y \to x \)) post-nonlinear model.  We again validate this subjective impression by considering the respective log-likelihoods:  The bound on \( L_{y \to x, \text{\subref{fig:alt_sun-1}}} \) is approximately \num{-2.30} for the post-nonlinear model.  The exact log-likelihood \( L_{y \to x, \text{\subref{fig:alt_sun-2}}} = -2.74 \) under the additive noise model is lower.

  \begin{figure}[b]
    \centering
    \def\labelwidth{10mm}
    \def\labelshift{-11mm}
    \begin{tikzpicture}
      \begin{groupplot}[
        group style = {
          rows = 1, columns = 2,
          horizontal sep = 1.5cm,
          vertical sep = 2cm,
        },
        width = 0.5\linewidth, height = 0.42\linewidth,
        ]

        \nextgroupplot[
        xlabel = \( x \), ylabel = \( y \),
        xmin = -3.5, xmax = 5.5, ymin = -4.5, ymax = 3.5]

        \addplot[only marks, mark size=0.75pt, myblue, opacity=0.3, draw opacity=0] table[x index = 1, y index = 0] {plot_data/alt_sun_dataset.txt};
        \addplot[only marks, mark size=0.75pt, myred, opacity=0.3, draw opacity=0] table[x index = 1, y index = 0] {plot_data/alt_sun_dataset_replication.txt};
        
        \nextgroupplot[
        xlabel = \( x \), ylabel = \( y \),
        xmin = -3.5, xmax = 5.5, ymin = -4.5, ymax = 3.5]

        \addplot[only marks, mark size=0.75pt, myblue, opacity=0.3, draw opacity=0] table[x index = 1, y index = 0] {plot_data/alt_sun_dataset.txt};
        \addplot[only marks, mark size=0.75pt, myred, opacity=0.3, draw opacity=0] table[x index = 1, y index = 0] {plot_data/alt_sun_dataset_replication_gauss.txt};
      \end{groupplot}

      \node[text width = \labelwidth] at ([yshift=\labelshift] group c1r1.south) {\subcaption{\label{fig:alt_sun-1}}};
      \node[text width = \labelwidth] at ([yshift=\labelshift] group c2r1.south) {\subcaption{\label{fig:alt_sun-2}}};
    \end{tikzpicture}

    \caption[Investigation of pair~4 (altitude~\( \to \)~sunshine duration)]{Investigation of pair~4 (altitude~\( \to \)~sunshine duration): The anti-causal direction follows a post-nonlinear model, but not an additive noise model.  This can be seen if comparing the replication of the original dataset (small blue dots) by generating samples from the estimated post-nonlinear model (small red dots in panel~\subref{fig:alt_sun-1}) and the samples taken from the estimated additive noise model (small red dots in panel~\subref{fig:alt_sun-2}).
    }
    \label{fig:alt_sun}
  \end{figure}
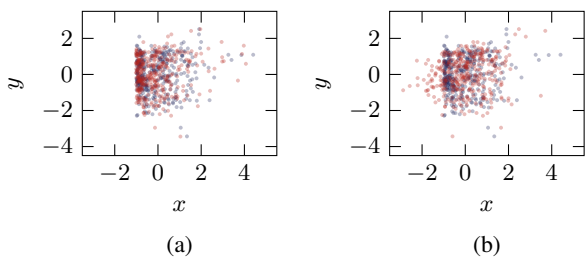

  Nevertheless, taking Occam's Razor into consideration, we would prefer the simpler additive noise-model for \( x \to y \) over the post-nonlinear model for \( y \to x \).  Exploiting this reasoning, a possible two-step procedure could be to first test if the data can be described by an additive noise model and only consider the post-nonlinear model if not.  The difficulty lies in defining a plausible threshold on the \( p \)-value which is required for data to follow an additive noise model.

\item \emph{Overall results on the whole database:} \nopagebreak
  
  Our method is constructed to cope with data which are one-dimensional in \( x \) and \( y \).  \num{102} of \num{108} pairs of the database satisfy this condition.  We evaluated the test statistic of \ac{HSIC} and the estimated log-likelihoods \eqref{eq:L_xy}, \eqref{eq:L_yx} for \num{25} independent runs, comparing our method with the classic approach based on the additive noise model.  To exclude outliers inside the \num{25} samples (which occur due to the more challenging optimisation problem compared to the simpler additive noise model), we use the median of the \num{25} samples for the final model selection.

  Choosing the \ac{HSIC} test statistic as the model selection criterion, our method achieves an accuracy of \SI{68.8}{\percent} compared to \SI{59.8}{\percent} for the additive noise model approach.  The authors of \cite{mooijDistinguishingCauseEffect2016} suggest to weight each pair of the database as some pairs are correlated to each other.  Based on this weighting, our accuracy decreases to \SI{59.6}{\percent} compared to \SI{62.4}{\percent} for the additive noise model approach.  The median test statistics for each pair are reported in Table~\ref{tab:results-tuebingen}.  Overall, there are \num{24}~pairs which are similar to the one presented in \ref{item:1}, \ie  where our method correctly identifies cause and effect in contrast to the additive noise approach.  On the contrary, \num{15}~pairs behave conversely, as the pair introduced in \ref{item:2}.

  Following the reasoning in \ref{item:2}, we combine our approach with the additive noise model approach:  Whenever one of the \( p \)-values we obtain after estimating additive noise models for \( x \to y \) and \( y \to x \) exceeds some threshold \( t \), we conclude that the additive noise model adequately describes the data.  In this case, we perform causal discovery by means of the additive noise model.  Otherwise, the more expressive post-nonlinear model is considered.  Choosing \( t \) appropriately is demanding and requires further investigation.  We leave it at evaluating how the (weighted) accuracy behaves for different thresholds.  The results are plotted in Figure~\ref{fig:combining-approaches}.  Choosing \( t \) within \SI{1}{\percent} and \SI{5}{\percent}, an accuracy of up to \SI{74.5}{\percent} (weighted accuracy of \SI{67.7}{\percent}) can be achieved which is an improvement of more than five percent points.

  Model selection based on \( L_{x\to y} \) and \( L_{y \to x} \), the lower bounds on the log-likelihoods, fails with respect to causal discovery.  The achieved weighted accuracy of \SI{53.3}{\percent} is not significantly higher than if deciding by chance.  We assume this to be caused by bounding the log-likelihood.  We therefore suggest to only consider \ac{HSIC} as a suitable criterion for bivariate causal discovery when using our estimation method for post-nonlinear models.

  \begin{figure}
    \centering
    \begin{tikzpicture}
      \begin{axis}[
        xmode = log,
        xmin = 3e-11, xmax = 3e0, ymin = 0.58, ymax = 0.78,
        xlabel = {threshold \( t \) on \( p \)-value}, ylabel = {accuracy},
        width = 0.9\linewidth, height = 0.6\linewidth,
        legend pos = north west
        ]


        \addplot[myblue] table {plot_data/combining_anm_pnm_acc.txt};
        \addlegendentry{accuracy}
        \addplot[myred] table {plot_data/combining_anm_pnm_acc_weighted.txt};
        \addlegendentry{weighted accuracy}
      \end{axis}
    \end{tikzpicture}
    \caption[Combining the additive noise model and the post-nonlinear model]{Combining the additive noise model and the post-nonlinear model.  The extreme case \( t = 1 \) corresponds to always deciding based on the post-nonlinear model.  Conversely, \( t = 0 \) only considers the additive noise model.  Using a threshold within \( [\SI{1}{\percent}, \SI{5}{\percent}] \), the accuracy can be improved by approximately five percent points.}
    \label{fig:combining-approaches}
  \end{figure}
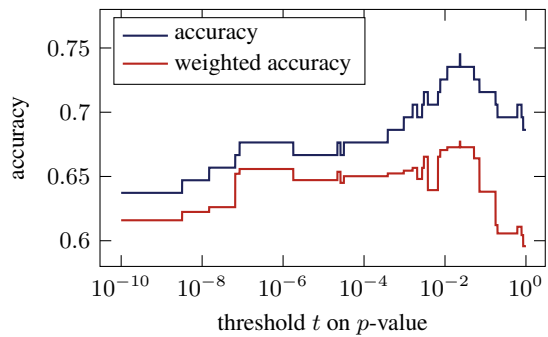

\item \emph{Missing quantitative comparison to \cite{zhangEstimationFunctionalCausal2016}'s method:}

  What is still lacking, is a comparison to the method our approach comes closest to: \cite{zhangEstimationFunctionalCausal2016}.  The authors report an (unweighted) accuracy of \SI{76}{\percent} which seems to be noticeably higher than our result.  However, at the time when their paper was published, the whole Tübingen database consisted of only \num{77} pairs.  Unfortunately, we could not examine \cite{zhangEstimationFunctionalCausal2016}'s work on today's full database as the provided algorithm is computationally time-consuming.\footnote{Moreover, we could not find out on which \num{77}~pairs of the database their reported accuracy rests on.  The oldest publicly available version of the Tübingen Database (released in June 2014) already contains \num{86}~pairs of which \num{81} are one-dimensional in \( x \) and \( y \).}  The estimation of each post-nonlinear model takes roughly ten~hours.  In comparison, our estimation method takes less than one minute, using the same hardware environment.
\end{enumerate}

\section{Conclusion}
\label{sec:conclusion}

In this paper, we presented a new regression technique which can be used to model post-nonlinear causal mechanisms.  The method has proven that it is capable of modelling artificially generated data which follow post-nonlinear models.  With the established procedures to distinguish cause and effect, we could identify cause and effect for identifiable structural causal models.

However, inference on practical data, where the underlying type of structural causal models is hidden, is still a demanding problem:  the post-nonlinear model is sometimes too expressive.  Following Occam's Razor, causal discovery benefits from considering the simpler additive noise model in such cases.  By introducing an intuitive criterion for when to apply which of the two models, we could improve the accuracy on the Tübingen database by more than five percent points.

\clearpage

\begin{table*}[h]
  \caption[Results of causal discovery on the Tübingen database]{Results of causal discovery on the Tübingen database.  Comparison of the \ac{HSIC} test statistic for the causal (\( x \to y \)) and the anti-causal (\( y \to x \)) direction. (Note that we always relabelled the cause-effect pairs such that \( x \) is the cause of \( y \).  In the original database, \( x \) and \( y \) are interchanged for some pairs.)  Our method based on the post-nonlinear model (\ac{PNM}) correctly identifies \num{70}~pairs compared to \num{61}~pairs when using exact Gaussian process regression with the additive noise model (\ac{ANM}) \cite{hoyerNonlinearCausalDiscovery2009, mooijDistinguishingCauseEffect2016}.  Those cause-effect pairs where causal discovery succeeded are shaded.}

  \centering
  \small
  \begin{tabular}{S[table-format=3.0]*{4}{S[table-format=2.3]}}
    \toprule
    & \multicolumn{2}{c}{\ac{PNM}} & \multicolumn{2}{c}{\ac{ANM}} \\
    \cmidrule(lr){2-3} \cmidrule(lr){4-5}
    {pair} & {\( x \to y \)} & {\( y \to x \)} & {\( x \to y \)} & {\( y \to x \)} \\
    \midrule
    1 & {\cellcolor{myred!20}} 0.514 & {\cellcolor{myred!20}} 1.221 & {\cellcolor{myblue!20}} 0.662 & {\cellcolor{myblue!20}} 2.292 \\
    2 & {\cellcolor{myred!20}} 0.360 & {\cellcolor{myred!20}} 1.331 & 2.730 & 2.074 \\
    3 & 0.416 & 0.285 & 0.789 & 0.753 \\
    4 & 0.927 & 0.674 & {\cellcolor{myblue!20}} 0.691 & {\cellcolor{myblue!20}} 1.036 \\
    5 & {\cellcolor{myred!20}} 0.383 & {\cellcolor{myred!20}} 0.656 & {\cellcolor{myblue!20}} 0.498 & {\cellcolor{myblue!20}} 2.469 \\
    6 & {\cellcolor{myred!20}} 0.256 & {\cellcolor{myred!20}} 0.494 & 6.219 & 2.670 \\
    7 & {\cellcolor{myred!20}} 0.380 & {\cellcolor{myred!20}} 1.098 & {\cellcolor{myblue!20}} 0.906 & {\cellcolor{myblue!20}} 2.962 \\
    8 & {\cellcolor{myred!20}} 0.229 & {\cellcolor{myred!20}} 0.583 & {\cellcolor{myblue!20}} 1.074 & {\cellcolor{myblue!20}} 2.602 \\
    9 & {\cellcolor{myred!20}} 0.363 & {\cellcolor{myred!20}} 0.833 & 4.863 & 1.574 \\
    10 & {\cellcolor{myred!20}} 0.543 & {\cellcolor{myred!20}} 0.557 & 5.747 & 1.717 \\
    11 & {\cellcolor{myred!20}} 0.271 & {\cellcolor{myred!20}} 0.620 & 6.109 & 3.023 \\
    12 & 0.446 & 0.359 & 8.528 & 3.245 \\
    13 & {\cellcolor{myred!20}} 0.172 & {\cellcolor{myred!20}} 0.459 & 4.661 & 3.485 \\
    14 & {\cellcolor{myred!20}} 0.199 & {\cellcolor{myred!20}} 0.392 & 2.496 & 2.284 \\
    15 & {\cellcolor{myred!20}} 0.388 & {\cellcolor{myred!20}} 0.432 & 4.007 & 1.370 \\
    16 & {\cellcolor{myred!20}} 0.251 & {\cellcolor{myred!20}} 0.263 & {\cellcolor{myblue!20}} 1.077 & {\cellcolor{myblue!20}} 2.014 \\
    17 & 0.821 & 0.382 & 37.234 & 0.997 \\
    18 & 0.517 & 0.278 & {\cellcolor{myblue!20}} 1.660 & {\cellcolor{myblue!20}} 4.842 \\
    19 & {\cellcolor{myred!20}} 0.137 & {\cellcolor{myred!20}} 0.239 & {\cellcolor{myblue!20}} 0.243 & {\cellcolor{myblue!20}} 1.711 \\
    20 & {\cellcolor{myred!20}} 2.090 & {\cellcolor{myred!20}} 2.442 & 3.577 & 2.279 \\
    21 & {\cellcolor{myred!20}} 0.387 & {\cellcolor{myred!20}} 0.860 & 1.616 & 0.645 \\
    22 & 0.663 & 0.602 & {\cellcolor{myblue!20}} 0.215 & {\cellcolor{myblue!20}} 0.594 \\
    23 & {\cellcolor{myred!20}} 0.728 & {\cellcolor{myred!20}} 0.854 & {\cellcolor{myblue!20}} 0.652 & {\cellcolor{myblue!20}} 1.429 \\
    24 & {\cellcolor{myred!20}} 0.545 & {\cellcolor{myred!20}} 0.617 & {\cellcolor{myblue!20}} 0.716 & {\cellcolor{myblue!20}} 0.772 \\
    25 & {\cellcolor{myred!20}} 0.413 & {\cellcolor{myred!20}} 0.504 & {\cellcolor{myblue!20}} 0.838 & {\cellcolor{myblue!20}} 1.706 \\
    26 & {\cellcolor{myred!20}} 0.525 & {\cellcolor{myred!20}} 1.632 & {\cellcolor{myblue!20}} 0.878 & {\cellcolor{myblue!20}} 2.731 \\
    27 & {\cellcolor{myred!20}} 0.821 & {\cellcolor{myred!20}} 0.847 & {\cellcolor{myblue!20}} 3.101 & {\cellcolor{myblue!20}} 5.314 \\
    28 & {\cellcolor{myred!20}} 0.557 & {\cellcolor{myred!20}} 1.182 & {\cellcolor{myblue!20}} 1.751 & {\cellcolor{myblue!20}} 4.421 \\
    29 & {\cellcolor{myred!20}} 0.311 & {\cellcolor{myred!20}} 2.644 & {\cellcolor{myblue!20}} 0.779 & {\cellcolor{myblue!20}} 4.015 \\
    30 & {\cellcolor{myred!20}} 0.639 & {\cellcolor{myred!20}} 1.041 & {\cellcolor{myblue!20}} 0.410 & {\cellcolor{myblue!20}} 1.100 \\
    31 & {\cellcolor{myred!20}} 0.453 & {\cellcolor{myred!20}} 0.823 & {\cellcolor{myblue!20}} 0.644 & {\cellcolor{myblue!20}} 1.489 \\
    32 & {\cellcolor{myred!20}} 0.232 & {\cellcolor{myred!20}} 3.039 & {\cellcolor{myblue!20}} 0.912 & {\cellcolor{myblue!20}} 28.474 \\
    33 & {\cellcolor{myred!20}} 0.443 & {\cellcolor{myred!20}} 1.298 & {\cellcolor{myblue!20}} 0.264 & {\cellcolor{myblue!20}} 3.095 \\
    34 & {\cellcolor{myred!20}} 0.303 & {\cellcolor{myred!20}} 0.425 & {\cellcolor{myblue!20}} 0.300 & {\cellcolor{myblue!20}} 1.656 \\
    35 & {\cellcolor{myred!20}} 0.632 & {\cellcolor{myred!20}} 0.873 & 1.343 & 1.223 \\
    36 & {\cellcolor{myred!20}} 0.367 & {\cellcolor{myred!20}} 0.664 & {\cellcolor{myblue!20}} 0.740 & {\cellcolor{myblue!20}} 2.250 \\
    37 & {\cellcolor{myred!20}} 0.409 & {\cellcolor{myred!20}} 1.337 & {\cellcolor{myblue!20}} 2.374 & {\cellcolor{myblue!20}} 3.643 \\
    38 & {\cellcolor{myred!20}} 0.423 & {\cellcolor{myred!20}} 0.838 & {\cellcolor{myblue!20}} 0.447 & {\cellcolor{myblue!20}} 1.168 \\
    39 & {\cellcolor{myred!20}} 0.241 & {\cellcolor{myred!20}} 0.806 & {\cellcolor{myblue!20}} 0.323 & {\cellcolor{myblue!20}} 2.281 \\
    40 & {\cellcolor{myred!20}} 0.208 & {\cellcolor{myred!20}} 0.690 & {\cellcolor{myblue!20}} 0.290 & {\cellcolor{myblue!20}} 5.401 \\
    41 & {\cellcolor{myred!20}} 0.320 & {\cellcolor{myred!20}} 0.598 & {\cellcolor{myblue!20}} 1.118 & {\cellcolor{myblue!20}} 3.389 \\
    42 & {\cellcolor{myred!20}} 0.206 & {\cellcolor{myred!20}} 4.696 & {\cellcolor{myblue!20}} 0.338 & {\cellcolor{myblue!20}} 33.150 \\
    43 & {\cellcolor{myred!20}} 0.240 & {\cellcolor{myred!20}} 0.308 & 8.944 & 8.896 \\
    44 & {\cellcolor{myred!20}} 1.188 & {\cellcolor{myred!20}} 1.417 & 1.723 & 1.235 \\
    45 & {\cellcolor{myred!20}} 0.809 & {\cellcolor{myred!20}} 1.022 & {\cellcolor{myblue!20}} 2.069 & {\cellcolor{myblue!20}} 4.043 \\
    46 & {\cellcolor{myred!20}} 0.493 & {\cellcolor{myred!20}} 0.895 & {\cellcolor{myblue!20}} 3.545 & {\cellcolor{myblue!20}} 5.744 \\
    47 & {\cellcolor{myred!20}} 0.222 & {\cellcolor{myred!20}} 1.637 & {\cellcolor{myblue!20}} 0.620 & {\cellcolor{myblue!20}} 7.832 \\
    48 & 0.843 & 0.781 & {\cellcolor{myblue!20}} 0.883 & {\cellcolor{myblue!20}} 2.108 \\
    49 & 0.568 & 0.215 & {\cellcolor{myblue!20}} 0.694 & {\cellcolor{myblue!20}} 1.684 \\
    50 & 1.253 & 0.521 & {\cellcolor{myblue!20}} 1.563 & {\cellcolor{myblue!20}} 1.776 \\
    51 & 1.195 & 0.935 & 2.026 & 1.273 \\
    \bottomrule
  \end{tabular}
  \qquad\qquad
  \begin{tabular}{S[table-format=3.0]*{4}{S[table-format=2.3]}}
    \toprule
    & \multicolumn{2}{c}{\ac{PNM}} & \multicolumn{2}{c}{\ac{ANM}} \\
    \cmidrule(lr){2-3} \cmidrule(lr){4-5}
    {pair} & {\( x \to y \)} & {\( y \to x \)} & {\( x \to y \)} & {\( y \to x \)} \\
    \midrule
    56 & 0.490 & 0.398 & 3.484 & 1.138 \\
    57 & {\cellcolor{myred!20}} 0.291 & {\cellcolor{myred!20}} 0.325 & 2.761 & 0.967 \\
    58 & {\cellcolor{myred!20}} 0.343 & {\cellcolor{myred!20}} 0.434 & 2.548 & 0.923 \\
    59 & 0.464 & 0.301 & 2.832 & 1.023 \\
    60 & 0.525 & 0.353 & 2.151 & 1.028 \\
    61 & {\cellcolor{myred!20}} 0.268 & {\cellcolor{myred!20}} 0.334 & 1.762 & 0.733 \\
    62 & {\cellcolor{myred!20}} 0.363 & {\cellcolor{myred!20}} 0.479 & 1.669 & 1.002 \\
    63 & 0.547 & 0.391 & 2.272 & 1.118 \\
    64 & {\cellcolor{myred!20}} 0.274 & {\cellcolor{myred!20}} 0.334 & {\cellcolor{myblue!20}} 3.131 & {\cellcolor{myblue!20}} 3.225 \\
    65 & 0.476 & 0.469 & {\cellcolor{myblue!20}} 1.032 & {\cellcolor{myblue!20}} 1.294 \\
    66 & 0.330 & 0.258 & {\cellcolor{myblue!20}} 0.822 & {\cellcolor{myblue!20}} 1.121 \\
    67 & {\cellcolor{myred!20}} 0.290 & {\cellcolor{myred!20}} 0.679 & {\cellcolor{myblue!20}} 1.372 & {\cellcolor{myblue!20}} 1.584 \\
    68 & 1.089 & 0.759 & 14.627 & 1.546 \\
    69 & 0.516 & 0.333 & 4.290 & 1.611 \\
    70 & 15.139 & 0.889 & 54.621 & 4.725 \\
    72 & {\cellcolor{myred!20}} 0.280 & {\cellcolor{myred!20}} 0.617 & {\cellcolor{myblue!20}} 0.761 & {\cellcolor{myblue!20}} 3.707 \\
    73 & 0.296 & 0.258 & {\cellcolor{myblue!20}} 19.834 & {\cellcolor{myblue!20}} 23.979 \\
    74 & {\cellcolor{myred!20}} 0.264 & {\cellcolor{myred!20}} 0.376 & {\cellcolor{myblue!20}} 2.387 & {\cellcolor{myblue!20}} 3.632 \\
    75 & {\cellcolor{myred!20}} 0.299 & {\cellcolor{myred!20}} 0.313 & 4.330 & 4.195 \\
    76 & {\cellcolor{myred!20}} 0.231 & {\cellcolor{myred!20}} 0.493 & {\cellcolor{myblue!20}} 0.381 & {\cellcolor{myblue!20}} 1.060 \\
    77 & {\cellcolor{myred!20}} 0.208 & {\cellcolor{myred!20}} 0.284 & {\cellcolor{myblue!20}} 1.623 & {\cellcolor{myblue!20}} 5.040 \\
    78 & 0.403 & 0.359 & {\cellcolor{myblue!20}} 2.153 & {\cellcolor{myblue!20}} 5.510 \\
    79 & 0.528 & 0.375 & {\cellcolor{myblue!20}} 2.095 & {\cellcolor{myblue!20}} 3.365 \\
    80 & {\cellcolor{myred!20}} 0.404 & {\cellcolor{myred!20}} 0.488 & {\cellcolor{myblue!20}} 1.988 & {\cellcolor{myblue!20}} 5.465 \\
    81 & {\cellcolor{myred!20}} 0.389 & {\cellcolor{myred!20}} 0.615 & {\cellcolor{myblue!20}} 0.657 & {\cellcolor{myblue!20}} 0.957 \\
    82 & {\cellcolor{myred!20}} 0.231 & {\cellcolor{myred!20}} 0.443 & 4.446 & 0.916 \\
    83 & {\cellcolor{myred!20}} 0.278 & {\cellcolor{myred!20}} 0.584 & 5.152 & 0.842 \\
    84 & 0.276 & 0.269 & {\cellcolor{myblue!20}} 0.433 & {\cellcolor{myblue!20}} 1.183 \\
    85 & {\cellcolor{myred!20}} 0.263 & {\cellcolor{myred!20}} 0.745 & {\cellcolor{myblue!20}} 0.516 & {\cellcolor{myblue!20}} 0.896 \\
    86 & {\cellcolor{myred!20}} 0.306 & {\cellcolor{myred!20}} 0.494 & {\cellcolor{myblue!20}} 0.944 & {\cellcolor{myblue!20}} 3.280 \\
    87 & 1.688 & 0.859 & 29.249 & 7.272 \\
    88 & {\cellcolor{myred!20}} 0.208 & {\cellcolor{myred!20}} 0.382 & 2.824 & 1.450 \\
    89 & 0.184 & 0.167 & {\cellcolor{myblue!20}} 0.223 & {\cellcolor{myblue!20}} 0.244 \\
    90 & 0.426 & 0.337 & {\cellcolor{myblue!20}} 0.437 & {\cellcolor{myblue!20}} 0.493 \\
    91 & {\cellcolor{myred!20}} 0.222 & {\cellcolor{myred!20}} 0.340 & {\cellcolor{myblue!20}} 1.223 & {\cellcolor{myblue!20}} 1.804 \\
    92 & 0.348 & 0.295 & 1.331 & 0.839 \\
    93 & 0.907 & 0.365 & 1.826 & 1.692 \\
    94 & {\cellcolor{myred!20}} 0.288 & {\cellcolor{myred!20}} 1.200 & {\cellcolor{myblue!20}} 1.234 & {\cellcolor{myblue!20}} 3.192 \\
    95 & {\cellcolor{myred!20}} 0.385 & {\cellcolor{myred!20}} 1.122 & {\cellcolor{myblue!20}} 0.754 & {\cellcolor{myblue!20}} 3.541 \\
    96 & {\cellcolor{myred!20}} 0.391 & {\cellcolor{myred!20}} 0.719 & {\cellcolor{myblue!20}} 1.644 & {\cellcolor{myblue!20}} 5.700 \\
    97 & {\cellcolor{myred!20}} 0.150 & {\cellcolor{myred!20}} 0.297 & 0.283 & 0.247 \\
    98 & {\cellcolor{myred!20}} 0.212 & {\cellcolor{myred!20}} 0.390 & 0.743 & 0.742 \\
    99 & {\cellcolor{myred!20}} 0.307 & {\cellcolor{myred!20}} 2.306 & {\cellcolor{myblue!20}} 0.871 & {\cellcolor{myblue!20}} 1.035 \\
    100 & {\cellcolor{myred!20}} 0.204 & {\cellcolor{myred!20}} 0.330 & {\cellcolor{myblue!20}} 0.689 & {\cellcolor{myblue!20}} 1.161 \\
    101 & {\cellcolor{myred!20}} 0.347 & {\cellcolor{myred!20}} 0.573 & {\cellcolor{myblue!20}} 0.811 & {\cellcolor{myblue!20}} 6.127 \\
    102 & {\cellcolor{myred!20}} 0.276 & {\cellcolor{myred!20}} 0.416 & {\cellcolor{myblue!20}} 0.450 & {\cellcolor{myblue!20}} 0.689 \\
    103 & 0.263 & 0.171 & 0.257 & 0.241 \\
    104 & 0.170 & 0.142 & 0.381 & 0.223 \\
    106 & 0.312 & 0.264 & {\cellcolor{myblue!20}} 0.529 & {\cellcolor{myblue!20}} 0.551 \\
    107 & 2.658 & 0.492 & 19.405 & 1.006 \\
    108 & {\cellcolor{myred!20}} 0.390 & {\cellcolor{myred!20}} 0.468 & 3.153 & 0.820 \\
    \bottomrule
  \end{tabular}

  \label{tab:results-tuebingen}
\end{table*}

\clearpage

{\small
  \bibliographystyle{ieee_fullname}
  \bibliography{refs}
}

\end{document}